\documentclass[sigconf]{acmart}
\usepackage{balance}
\AtBeginDocument{%
  }

\setcopyright{acmlicensed}
\copyrightyear{2018}
\acmYear{2018}
\acmDOI{XXXXXXX.XXXXXXX}

\acmConference[Conference acronym 'XX]{Make sure to enter the correct
  conference title from your rights confirmation emai}{June 03--05,
  2018}{Woodstock, NY}
\acmISBN{978-1-4503-XXXX-X/18/06}

\begin{document}

\title{Text Style Transfer with Machine Translation for Graphic Designs}


\author{Deergh Budhauria}
\email{deerg@adobe.com}
%\orcid{}
\affiliation{%
  \institution{Adobe}
  \city{Noida}
  \country{India}
}

\author{Rishav Agarwal}
\email{rishagar@adobe.com}
%\orcid{}
\affiliation{%
  \institution{Adobe}
  \city{Noida}
  \country{India}
}

\author{Sanyam Jain }
\email{sanyjain@adobe.com}
%\orcid{}
\affiliation{%
  \institution{Adobe}
  \city{Noida}
  \country{India}
}

\author{Tracy Holloway King}
\email{tking@adobe.com}
\orcid{0000-0002-7956-505X}

\affiliation{%
  \institution{Adobe}
  \city{San Jose}
  \state{California}
  \country{USA}
}

\renewcommand{\shortauthors}{Budhauria et al.}

\begin{abstract}
Globalization of graphic designs such as those used in marketing materials and magazines is increasingly important for communication to broad audiences. To accomplish this, the textual content in the graphic designs needs to be accurately translated and have the text styling preserved in order to fit visually into the design.  Preserving text styling requires high accuracy word alignment between the original and the translated text.
  The problem of word alignment between source and translated text is long known. The industry standards for extracting word alignments are defined by Giza++ and attention probabilities from neural machine translation (NMT) models. In this paper, we  explore three new methods to tackle the word alignment problem for transferring text styles from the source to the translated text. The proposed methods are developed on top of commercially available NMT and LLM translation technologies. They include: NMT with custom input and output tags for text styling; LLM with custom input and output tags; a hybrid with NMT for translation followed by an LLM with use of unigram mappings.  To analyze the performance of these solutions, their alignment results are compared with the results of an attention head approach to gauge their usability in graphic design applications. Interestingly, the  attention head strong baseline proves more accurate than the LLM or  NMT approach and on par with the hybrid NMT+LLM approach.  
\end{abstract}



 % \caption{Seattle Mariners at Spring Training, 2010.}

\maketitle

\section{	Introduction}

In an era defined by unprecedented global interconnectedness, the demand for effective cross-linguistic communication has never been greater. At the forefront of addressing this demand is the field of machine translation (MT). The requirements on MT capabilities are even greater in the domain of graphic design, where not only does text have to be translated but it has to preserve the text styling and each phrase has to be properly placed within the larger page design.  
\begin{figure}[htb]
 \includegraphics[width=2.6in]{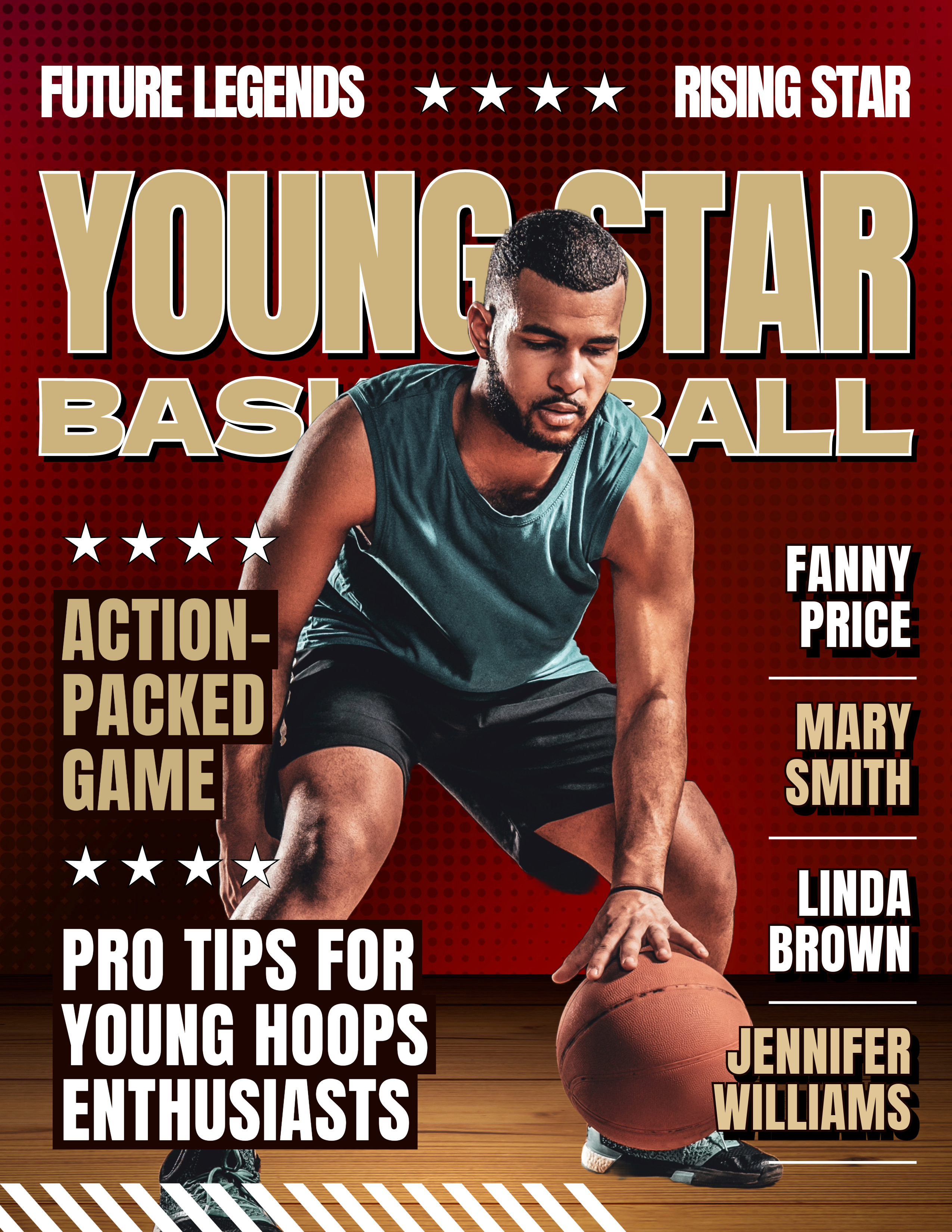} 
 
 \vspace*{1ex}
 
 \includegraphics[width=2.6in]{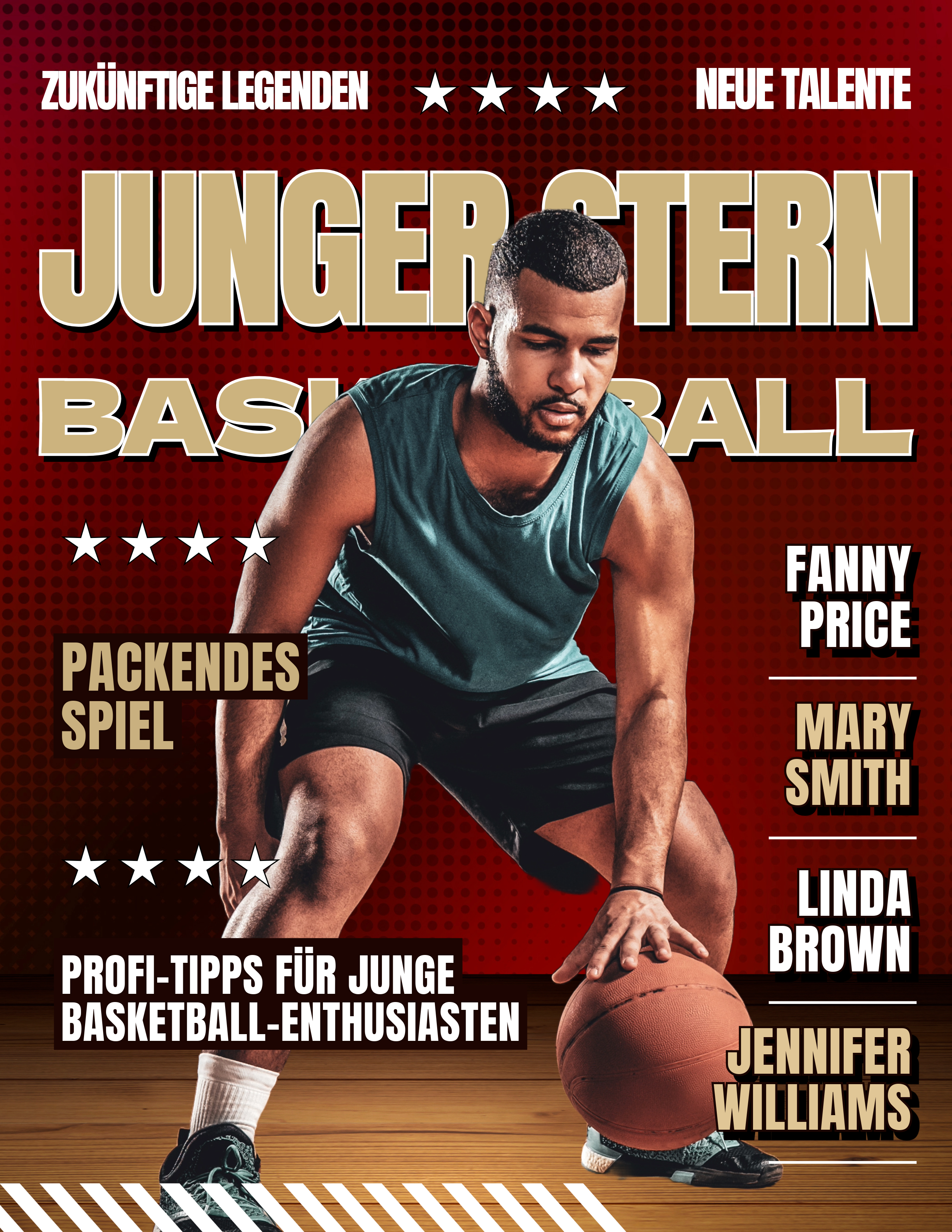}
\caption{ Example magazine-style pages: English top, German bottom. Different font sizes and colors are used for different text. Due to differences in phrasing, the text wrapping, which is a separate formatting parameter, differs across the languages.}
 \label{fig:exampledesign}
\end{figure}
Figure \ref{fig:exampledesign} shows the English and German version of a mock magazine cover which uses multiple font sizes and colors. %Figure \ref{fig:exampledesign} shows the English and French version of the Adobe.com home page which uses multiple font sizes as well as bold face. 
A graphic design needs to travel  globally with a native feel tuned to each region by having apt content language and styling aesthetics. In this competitive era, styling of text has become a must-have and  helps  businesses  stand out amongst their competitors. Multilingual designs create high demand for accurate translations. Producing accurate translations is  a difficult, much studied problem with major conferences (e.g.\ AMTA \cite{amta-2022-biennial}) and tracks in NLP conferences (e.g.\ ACL \cite{acl-2023-frontmatter}) devoted to it \citep{mohamedetal}. Adding the requirement of maintaining the styling information of the text  makes it even more difficult for MT to create an accurate  document. To provide an estimate for the demand for text styling in graphic design, approximately half of the exported documents in Adobe InDesign have text styles applied including the use of bold face, italics, color, and different fonts to differentiate parts of the text. 

This paper proposes two high accuracy methods to transfer text styles of a graphic design. Both approaches first obtain a high quality translation and then apply style transfer to the translated content (sections \ref{sec:attentionheads} and \ref{sec:hybrid}). We compare these to two other approaches: one using NMT with custom input and output tags (section \ref{sec:nmt}) and one using an LLM with custom input and output tags specified through prompt engineering (section \ref{sec:llm}).

The standard  solution to style transfer across translations is to create a word alignment map  between the source and  translated text. (As shown below, this is still a state-of-the-art solution even with recent advances to NMT and LLMs.) A word alignment map can be created using traditional techniques like Giza++ which is a statistical tool that generates alignment probabilities with Expectation Maximization (EM)  \citep{brownetal1993}.
Another solution is to leverage the attention from Neural Machine Translation (NMT). NMT constitutes the state of the art in MT, with the transformer model architecture outperforming other neural architectures   \citep{vaswanietal}.  NMT models use an attention mechanism to generate translations. The attention mechanism focuses closely on the context of the source text while generating the translation.  The attention probabilities can be used  to solve the problem of word alignment \citep{gargetal}   and the results are often at par (or better) than the statistical translation models which have been treated as a baseline for finding alignment probabilities for over a decade. 

In this work, we  use the transformer.wmt19.en-de model from Facebook’s fairseq toolkit as the base transformer model to generate machine translations as well  attention probabilities \citep{fairseq}. We  make use of the attention probabilities to map differently stylized words in the source text to appropriate words in the translated text and then transfer the styles between them. We treat these results as a strong baseline while evaluating three additional methods for transferring styles between source and the translated text. For all of the methods, we use commercially available NMT and LLM solutions. As the NMT- and LLM-only approaches perform worse than the attention head baseline, we propose a hybrid solution which uses NMT for high quality content translation, which is further fed to LLM to determine unigram mappings and do the style transfer.  We compare the word alignment maps from all four methods to gauge their accuracy for use in graphic designing products like Adobe InDesign and Adobe Illustrator  to help graphic designers more rapidly  translate and style text in their work. 

Contributions of this paper include:

\begin{itemize}
\item Proposal and comparison of four methods of text style transfer for machine translation. (1) Baseline method: NMT with attention head alignment; (2) NMT with custom input and output tags for text style; (3) LLM with custom input and output tags for text style; (4) Hybrid method of NMT for translation followed by LLM for text style transfer 
\item Internal and external evaluation determining that a 2-stage process of NMT for translation and then either attention heads or LLM with custom input and output tags for text style transfer outperforms other methods.
\item Combining translation and style transfer in one step (LLM or NMT) results in awkward translations or bad style transfer as the systems try to force the style transfer into a continuous phrase. 
\end{itemize}

\section{	Attention Head Strong Baseline}
\label{sec:attentionheads}

Current graphic design translations are done either by hiring human translators versed in different locales or by going out of the design application to do the translation of the plain text in other programs such as online translation services. This is  cumbersome for  designers and editors of graphic documents such as in Figure \ref{fig:exampledesign}. There are very limited solutions to this problem in existing products. Text editing and formatting application that do provide both machine translation and style transfer  often have a lossy style transfer thus messing up the design of the whole document. Improving this requires a way to transfer the text formatting from the source to the translation via word alignment.

Formally, alignment across two texts is defined as follows.
Given a sentence s$_{1}$$^{J}$ = s$_{1}$, $\ldots$, s$_{j}$, $\ldots$ s$_{J}$  in the source language and its translation t$_{1}$$^{I}$ = t$_{1}$, $\ldots$, t$_{i}$, $\ldots$ t$_{I}$ in the target language, an alignment A is defined as a subset of the Cartesian product of the word positions \cite{ochney}. 

\begin{equation}
A \subseteq {(j, i) : j = 1, \ldots, J; i = 1, \ldots, I}
\end{equation}

The lengths of the source and translated sentence may be, and often are, different, meaning that there are many-to-one and many-to-many alignments.  The word alignment task aims to find a discrete alignment representing a many-to-many mapping from the source words to their corresponding translations in the target sentence. In this section we consider three strategies for word alignment: Giza++,  cosine similarity of word vectors, Attention heads in NMT.

\subsection{Standard Word Alignment Strategies}

\textbf{Giza++} In the realm of word alignment research, Giza++ is a foundational tool for aligning words across bilingual text corpora. Its strength lies in the robust computation of alignment probabilities, which are essential for accurate word alignment. Giza++ implements a statistical alignment algorithm based on the Expectation-Maximization (EM) approach \citep{brownetal1993}. Through iterative refinement, Giza++ optimizes alignment probabilities between words in the source and target languages, maximizing the likelihood of observed parallel corpus data. Alignment probabilities in Giza++ are estimated by considering both lexical and positional information. The lexical component assigns probabilities based on translational equivalence and co-occurrence frequencies, while the positional component refines alignments by considering the relative positions of words in source and target sentences. %This dual approach enhances the accuracy of word alignments, accommodating various linguistic phenomena and bidirectional translation scenarios. Moreover, Giza++ offers parameterization flexibility, enabling customization of alignment models to suit specific language pairs and text genres. By providing reliable alignment probabilities, Giza++ contributes significantly to our understanding of bilingual lexical correspondences and facilitates the development of more precise alignment algorithms in natural language processing.
Unfortunately, since Giza++ generates alignment probabilities dynamically and generates alignment only for the words in its source and target corpus, it is unusable for graphic designs. This is because the limited text content in graphic designs hampers the ability of Giza++ to generate alignment probabilities. The lack of data for Giza++ to act on drastically reduces its accuracy thus making it unsuitable for this application.

\textbf{Cosine Similarity }
Cosine similarity can be used to measure the similarity between translated sentences, documents, or even entire corpora. By quantifying the similarity between translations based on their vector representations, cosine similarity provides a valuable metric for evaluating the quality and adequacy of translations generated by machine translation systems. We can calculate the cosine similarity between the source and translated words (i.e.\ the potentially aligned words for text style transfer)  by representing them as unit length vectors in a vector space. This vector space is the set of word embeddings for all words in the dictionary of the source and target language.

\begin{equation}
cosine\_similarity(A,B) =  \frac{(A\cdot{B})}{(\parallel A \parallel\parallel B\parallel)	}
\label{eq:cosinesim}
\end{equation}

\noindent where A$\cdot$B denotes the dot product of vectors A and B, and $\parallel$A$\parallel$ and $\parallel$B$\parallel$ represent the Euclidean norms of vectors A and B, respectively. Although the graphic design use case  operates on sentences and phrases (e.g.\ headlines and captions), unfortunately, this simple, brute force technique is inefficient and does not account for the many-to-one and many-to-many alignments found in translation scenarios. However,  we  use cosine similarity to filter the word alignments provided by the attention heads (next section).

\subsection{Attention Head Alignment Algorithm}

Instead of using Giza++ or cosine similarity which do not work well for word alignment in cross-lingual graphic design scenarios, we leverage the alignment maps created using attention probabilities as our strong baseline. In neural machine translation (NMT), attention mechanisms have emerged as a foundational component facilitating the alignment of source and target language sequences \cite{vaswanietal,luongetal}. Central to this mechanism are attention probabilities, which dynamically weigh the relevance of each token in the source sequence while generating a given token in the target sequence.

\begin{figure}[htb]
 \includegraphics[width=3in]{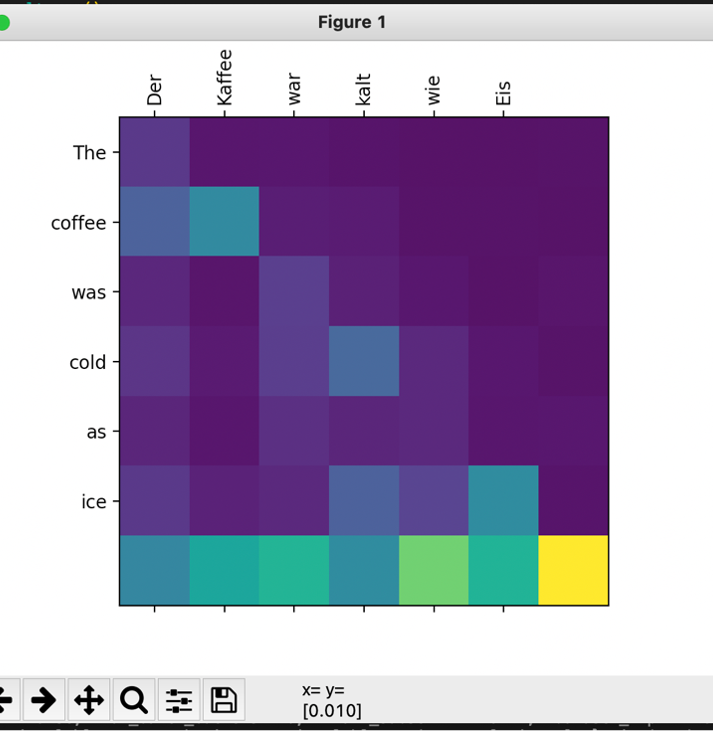}
\caption{ 2D matrix representing attention head values across English and German versions of the sentence \textit{The coffee was cold as ice.} In this translation pair, the words are in the same order and so the alignment should be along the diagonal.}
\label{fig:attention}
\end{figure}

Attention probabilities are computed via soft attention mechanisms, where the contextual representation of the decoder state interacts with the encoded representations of the source sequence through a compatibility function. Common compatibility functions include the dot product and multilayer perceptrons, which gauge the similarity between decoder and encoder states. Subsequently, a softmax operation is applied to normalize these weights, yielding attention probabilities that signify the salience of individual tokens in the source sequence \citep{bahdanauetal}.

The significance of attention probabilities for the task of text styling translation lies in their role in directing the focus of the translation process, allowing the model to selectively attend to pertinent information during decoding. This selective attention mechanism not only fosters improved alignment between source and target sequences but also enhances the coherence and fluency of generated translations by enabling the model to prioritize contextually relevant information \citep{luongetal}. Therefore we adopt this approach \citep{gargetal} as our  baseline.

We  used the  wmt19.en-de transformer model from Facebook’s fairseq toolkit as the base transformer model to generate machine translations and the attention probabilities required to create the attention map \citep{fairseq}.
To create the attention head model, we extracted the attention probabilities from the last layer of the wmt19.en-de model. This is a 2D matrix, where each row corresponds to a word of the source language sentence and each element of the row is the attention head probability of the words from the translated sentence (illustrated  in Figure \ref{fig:attention}). The higher the value of attention head, the higher  the correlation between the two words.

For each i$^{th}$ word in the set of specially styled words  from the source text, we compute the index of the top 3 attention probabilities from the corresponding row of the attention matrix. For each word corresponding to the top 3 indices, we find their cosine similarities (equation \ref{eq:cosinesim}) with the i$^{th}$ word. We empirically define a threshold value. If the cosine similarity value lies over this threshold, the word is added to the word alignment map and later styles are transferred to the translated text.

\begin{figure*}[htb]
    \includegraphics[width=5in]{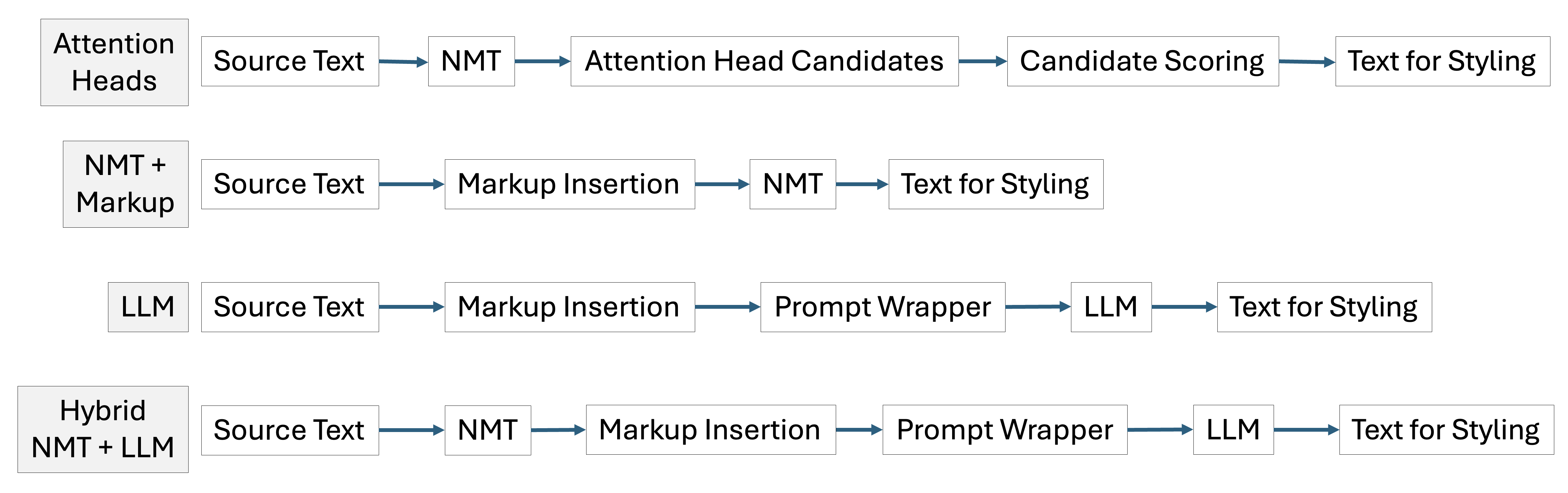} 
    \caption{Style preserving translation system architectures}
    \label{fig:systemdiagram}
\end{figure*}

\section{	Experiments}
To solve the problem of transferring styles from the source to the translated graphic design as a commercially-viable solution, we explored multiple methods. We looked at  the top commercially available options that provide machine translation: We selected the Microsoft translation API with its underlying NMT for our experiments \citep{MStrans}.\footnote{\url{https://azure.microsoft.com/en-us/products/ai-services/ai-translator}}\ We also needed a high quality commercially available LLM: We selected GPT 4 via Azure's OpenAI GPT for our experiments \citep{gpt}.\footnote{\url{https://azure.microsoft.com/en-us/products/ai-services/openai-service}}\  Using these models, we experimented with four  techniques to incorporate text style transfer. 

As discussed in section \ref{sec:attentionheads}, attention heads provide a promising method for determining text style placement.
However, none of the commercially available solutions that provide machine translations include attention probabilities: Instead, the attention heads have to be separately created as described in section \ref{sec:attentionheads} for our baseline.  We compared our baseline attention head approach with three approaches (Figure \ref{fig:systemdiagram}):

\begin{itemize}
    \item NMT with custom input and output tags for text style (section \ref{sec:nmt})
    \item LLM with custom input and output tags for text style (section \ref{sec:llm})
    \item Hybrid: NMT for translation followed by LLM with custom input and output tags for text style transfer (section \ref{sec:hybrid})
\end{itemize}

\subsection{	NMT (MS translation API)}
\label{sec:nmt}
The most widely used  commercial machine translation is Microsoft’s translation API. It provides translations to and from 100+ languages \citep{MStrans} with a variety of features that are accessed by adding  parameters to the service request. One such feature is built-in support for translating HTML pages. Only the text gets translated while the API response (the translated content) retains the HTML tags instead of   translating them.

We took advantage of this to transfer style during the translation. We created custom HTML tags for the styles. We use these  tags to enclose the text that is formatted differently. That is, we alter the input to the NMT system in order to translate the original text while leveraging the HTML tags for style information.  We pass this morphed input text as the request body and the response retains the custom style tags and places them at appropriate places in the translated content. This identifies the alignment maps of  styled words. Consider the example  in Table \ref{tab:mstranslation} where we translate an English sentence to German using the MS NMT API. Using the API response, we can appropriately transfer the styles between the source and the translated text.

\begin{table}[htb]
\begin{tabular}{l}
 \textbf{Source Sentence:} Job cuts have also soared \textit{nearly fivefold} so\\
 \hspace*{1em} far this year from a year ago.\\
 \textbf{API Request Body:} Job cuts have also soared $<$S1$>$nearly \\
 \hspace*{1em} fivefold$</$S1$>$ so far this year from a year ago.\\
 \textbf{API Response Translation:} Auch der Stellenabbau hat sich \\
 \hspace*{1em} in diesem Jahr im Vergleich zum Vorjahr $<$S1$>$fast \\
\hspace*{1em}  verfünffacht$</$S1$>$.
 \end{tabular}
    \caption{Example translated sentence with formating (italics on `nearly fivefold') from English to German via NMT with custom input and output tags using the MS NMT API.}
    \label{tab:mstranslation}
\end{table}

 \subsection{LLM (Azure OpenAI GPT)}
 \label{sec:llm}
Large Language Models (LLMs) have had a profound impact on  NLP applications (e.g.\ see \citep{moradi2024exploring} for an example overview). They offer a variety of NLP features, including machine translation. Here, we use GPT 4 via Azure's OpenAI GPT \citep{gpt} to generate translations of text that has style information appended to it. We use prompt engineering to ensure that the LLM   retains this styling information  while generating the translations. 
To do this, we created custom delimiters and enclosed the especially stylized text in these delimiters to encode the styling information along with the input text. The prompt fed to LLMs is shown in Table \ref{tab:azurellm_prompt}.

\begin{table}[htb]
\begin{tabular}{l}
\textbf{LLM System Prompt:}\\ 
You are efficient in language translation. We have a task where\\
we have English statements with some delimiters. \\
\#\#start\#\# -$>$ marks the start of a special style around this text. \\
\#\#end\#\# -$>$ marks the end of a special style around the text. \\
You have to perform the language translation keeping the info\\
of special styles intact for the
semantic meaning of the part of\\
the sentence. \\
Translate the prompts to German while retaining the styling\\
info for the part of sentences, 
learn from the example how\\
delimiters are used and converted in the translated text. 
\end{tabular}
\caption{Prompt for LLM (Azure OpenAI GPT) translation}
\label{tab:azurellm_prompt}
\end{table}

The response from the LLM contains the appropriate translations along with the delimiters enclosing the text upon which that style needs to be transferred. Consider the example in Table \ref{tab:azurellm_trans_ex}. In this example, the phrase {\em fell below 10 million in February} is contained in the custom delimiters so that the LLM can understand which is the differently styled text and can appropriately generate translations retaining the styling information.

\begin{table}[htb]
    \begin{tabular}{l}
\textbf{Input:} A Labor Department report this week also showed\\
\hspace*{1em} the number of available positions \#\#start\#\#fell below 10\\
\hspace*{1em} million in February \#\#end\#\# for the first time in nearly two\\
\hspace*{1em} years.\\[1em]
\textbf{Output:} Ein Bericht des Arbeitsministeriums in\\
\hspace*{1em} dieser Woche zeigte auch, dass die Anzahl der verfügbaren\\
\hspace*{1em} Stellen \#\#start\#\# im Februar \#\#end\#\# erstmals seit fast zwei\\
\hspace*{1em} Jahren \#\#start\#\#  unter 10 Millionen fiel \#\#end\#\#. 
   \end{tabular}
    \caption{Example English input to Azure OpenAI GPT with formatting start and stop positions (prompt prefix as in Table \ref{tab:azurellm_prompt} not shown) and the output German translation with formatting. Even though the German translation does not have the two phrases adjacent to one another, the text styling is correct.}
    \label{tab:azurellm_trans_ex}
\end{table}

\subsection{	Hybrid: NMT+LLM}
\label{sec:hybrid}
The translation quality of LLMs is worse than the accuracy of NMT machine translation models because of the generic nature of the data they are trained upon. To counter this, instead of asking an LLM to generate translations as in the direct LLM approach in section \ref{sec:llm}, we  propose a hybrid method. We first translate the text with an NMT. We then provide the LLM with both the input and output translated text, including the input styling, and  ask the LLM to transfer styling from the source to the translated text. The prompt provided to the LLM is in Table \ref{tab:hybridprompt}.

\begin{table}[htb]
    \begin{tabular}{l}
\textbf{System Prompt:} You are an expert in multiple languages.\\
You have to provide unigram mappings of the input words. \\
The input also contains source and target sentence in English\\
and German respectively. The unigram mappings to be provided\\
by you should be words from target sentence.
    \end{tabular}
    \caption{Prompt for the hybrid system}
    \label{tab:hybridprompt}
\end{table}

\begin{table}[htb]
    \begin{tabular}{l}
\textbf{Input:} \\
Source: A Labor Department report this week also showed the\\
\hspace*{1em} number of available positions fell below 10 million in February\\
\hspace*{1em} for the first time in nearly two years. \\
Target: Ein Bericht des Arbeitsministeriums in dieser Woche\\
\hspace*{1em}zeigte auch, dass die Anzahl der verfügbaren Stellen im Februar\\
\hspace*{1em} erstmals seit fast zwei Jahren unter 10 Millionen fiel. \\
Unigram: \{`fell', `below', `10', `million', `in', `February'\}\\[1em]
\textbf{Response}: maps: \{`fiel', `unter', `10', `Millionen', `im', `Februar'\}
    \end{tabular}
    \caption{Example input (prompt prefix as in Table \ref{tab:hybridprompt} not shown) and response for the hybrid system where the output is the German words to be stylized.}
    \label{tab:hybrid}
\end{table}

\begin{table*}[htb]
\begin{tabular}{lll|c|cc|cc|cc|cc}
\# &Text Style&	phrase(s)&	Eng.  &\multicolumn{2}{|c|}{Attention} & \multicolumn{2}{|c|}{NMT} & \multicolumn{2}{|c|}{LLM} & \multicolumn{2}{|c}{Hybrid}\\
& & & Cont.& Cont. & OK &Cont. & OK &Cont. & OK &Cont. & OK \\
											\hline
1 & italics+bold & fell below 10 million in February & y & n & \checkmark & n & X & y & \checkmark & n & \checkmark\\
2 & hyperlinks & nearly fivefold & y & y & \checkmark & y & \checkmark & y & \checkmark & y & \checkmark\\
3 & underline  & Speaker Kevin McCarthy in Los Angeles & y & n & \checkmark & n & X & y & \checkmark & n & \checkmark\\
4 & italics & familiar with the committee's & y & n & \checkmark & n & X & n & X & n & X\\
5 & highlight  & went viral & y & y & \checkmark & y & \checkmark & y & \checkmark & y &\checkmark \\
6 & highlight  & varying & y & y & \checkmark & y & \checkmark & y & \checkmark & y & \checkmark\\
7 & bold+hyperlink & Stassi Schroeder, Jax Taylor, Kristen  & y & y & \checkmark & y & \checkmark & y & \checkmark & y & \checkmark\\
& & Doute, Katie Maloney, Scheana Shay &&&&&&&&\\
8 & bold+hyperlink & Kristin Cavallari Sarah Michelle Gellar & n & n & \checkmark & n & \checkmark & n & \checkmark & n & \checkmark\\
9 & hyperlinks & 10th wedding anniversary & y & y & \checkmark & y & \checkmark & y & \checkmark & y & \checkmark \\
10 & underline & call following the discussion & y & y & \checkmark & y & \checkmark & y& \checkmark & y& \checkmark\\
\end{tabular}
\caption{Analysis of 10 real-world sentences with typographic text styling. \\
Cont. indicates whether the phrase was contiguous in the English and the German translation: y = yes, n = no. OK indicates correct typographical text styling or not: \checkmark = correct; X = incorrect}
\label{tab:errors}
\end{table*}

Consider the  example in Table \ref{tab:hybrid}. The LLM response contains the unigram mappings of the set of words fed as input. This unigram map is used to apply style transfer after. This way, the LLM returns the unigram maps that we use to generate the alignment maps to apply the style transfer. 

\section{	Qualitative analysis}
\label{sec:qualanalysis}

This section describes two analyzes: A smaller internal qualitative analysis of all four methods and a larger external analysis of the LLM and NMT methods. Insights  as to the strengths and weaknesses of each method are provided at the end of this section.

 \subsection{Small Scale Internal Analysis}
To analyze the accuracy of the solutions, we  compared the word alignment maps created by the three LLM and NMT approaches with the word alignment maps created by the attention head approach. We performed a qualitative analysis for sentences with assorted styles (e.g.\ bold, italics,  underline,  hyperlinks). 

Table \ref{tab:exampleonetwothree} shows two of these with the text style shown as italics. For each source, we  provide the results after applying the style transfers. The translations  differ slightly between systems, due in part to the freer word order in German compared to English.
Both examples show a similar pattern which highlights the main issues with the LLM and NMT approaches compared to the hybrid and attention head approaches. The attention head and hybrid approaches have correct text styling and natural sounding translations. Of particular interest is that fact that the natural German translations split the stylized English text into two non-contiguous phrases.  The LLM approach has correct text styling but  forces a German translation where the stylized text is contiguous, as it was in the English, which is less natural sounding. The NMT approach shows two different failures: in example 1, additional text \textit{am Mittwoch} is stylized; in example 2, the text \textit{im Februar} has failed to be stylized.

We  compiled a broader set of examples to determine whether these observations generalize. Ten of these are summarized in Table \ref{tab:errors}. We  make the following observations. 
The type of text styling does not affect any of the algorithms, i.e.\ they perform equally well with all types of styling.
Of greatest interest is the fact that the strong baseline of using the attention heads out-performed all the other methods (NMT+content tags, LLM+style tags, hybrid NMT+LLM). The hybrid approach has the same high quality translation as the attention head and NMT approach since they share the same underlying translation engine (Figure \ref{fig:systemdiagram}) and  leverages the improved text styling of the LLM (see Table \ref{tab:vendoreval} and the discussion in the next section).

\subsection{External Analysis}
We conducted an external analysis to get expert opinions on real world graphic design documents,  focusing on the comparison of the NMT  and LLM  techniques. We prepared a dataset of 35 graphic design documents and corresponding generated translated German versions from the NMT and LLM approaches, including retaining the styling of the document according to the systems' recommendations. The documents were in 3 categories: Short (1--2 Pages), Medium (5--15 Pages) and Long documents (30+ pages). Each document had multiple pages with multiple paragraphs. This results in  $\sim$2500 sentences. Translation experts  evaluated the translated versions  on  aspects of content translation (with internal factors of adequacy and fluency) and style transfer (on a scale of 1-5). Table \ref{tab:vendoreval} captures the assessment results. Long form documents were  only analyzed for content translation accuracy since from an initial analysis of the short and medium documents, it became evident that the style transfer depends on the content being appropriately translated. 

\begin{table}[htb]
    \begin{tabular}{lllll}
& \multicolumn{2}{c}{NMT} & \multicolumn{2}{c}{LLM}\\
\hline
Doc Type  & Trans. & Style & Trans. & Style\\
Short (15) & 2.53 & 4.8 & 2.53 & 4.87\\
Medium (8) & 2 & 3.17 & 2.25 & 3.57\\
Long (11) & 2.72 & & 1.64 & \\
\end{tabular}
    \caption{Results of the external evaluation for the NMT and LLM approaches on $\sim$2500 sentences. Trans.\ indicates the content translation score on a scale of 1 to 5. Style indicates the style transfer score on a scale of 1 to 5.}
    \label{tab:vendoreval}
\end{table}

Based on the large scale external evaluation,  NMT  outperforms the LLM in terms of content translation, whereas for style transfer of the  LLM is marginally better. However, as we observed in the smaller scale internal evaluations, both these techniques fall behind the attention head and  hybrid NMT+LLM method.

 \begin{table*}[htb]
    \begin{tabular}{|l|l|}
\hline
\multicolumn{2}{|l|}{Example 1: Attention heads and hybrid have correct text styling. NMT incorrectly styles \textit{am Mittwoch}.}\\
\multicolumn{2}{|l|}{\hspace*{2em}LLM has correct styling but awkward word order.}\\\hline
Original Content &	The three-day drills, announced the day after Tsai returned from the United States, had been widely\\ 
& expected after Beijing condemned her Wednesday meeting with \textit{Speaker Kevin McCarthy in Los Angeles}. \\[1ex]
Attention heads &	Die dreitägigen Übungen, die am Tag nach Tsais Rückkehr aus den USA angekündigt wurden, waren \\
& algemein erwartet worden, nachdem Peking ihr Treffen mit dem \textit{Sprecher Kevin McCarthy} am Mittwoch \\
&\textit{in Los Angeles} verurteilt hatte.  \\[1ex]
NMT (MS API  &	Die dreitägigen Übungen, die am Tag nach Tsais Rückkehr aus den Vereinigten Staaten angekündigt wurden, \\
with custom tags ) &waren weithin erwartet worden, nachdem Peking ihr Treffen mit \textit{dem Sprecher des Repräsentantenhauses,} \\
&\textit{Kevin McCarthy, am Mittwoch in Los Angeles} verurteilt hatte. \\[1ex]
LLM (Azure OpenAI  &	Die dreitägigen Übungen, die am Tag nach Tsais Rückkehr aus den Vereinigten Staaten angekündigt wurden,\\
with custom delimiters)&  waren nach Pekings Verurteilung ihres Treffens am Mittwoch mit \textit{dem Sprecher Kevin McCarthy in Los} \\
&\textit{Angeles} weitgehend erwartet worden.  \\[1ex]
Hybrid &Die dreitägigen Übungen, die am Tag nach Tsais Rückkehr aus den USA angekündigt wurden, waren \\
&algemein erwartet worden, nachdem Peking ihr Treffen mit dem \textit{Sprecher Kevin McCarthy} am Mittwoch  \\
&\textit{in Los Angeles} verurteilt hatte.\\
\hline\hline
\multicolumn{2}{|l|}{Example 2: Attention heads and hybrid have correct text styling. NMT fails to style \textit{im Februar}.}\\
\multicolumn{2}{|l|}{\hspace*{2em}LLM has correct styling but awkward word order.}\\\hline
Original Content &	A Labor Department report this week also showed the number of available positions \textit{fell below 10 million}\\
&\textit{in February}  for the first time in nearly two years.\\[1ex]
Attention heads &	Ein Bericht des Arbeitsministeriums in dieser Woche zeigte zudem, dass die Zahl der verfügbaren Stellen \\
&\textit{im Februar} erstmals seit fast zwei Jahren \textit{unter 10 Millionen fiel}. \\[1ex]
NMT (MS API   &	Ein Bericht des Arbeitsministeriums in dieser Woche zeigte auch, dass die Zahl der verfügbaren Stellen  \\
with custom tags )& im Februar zum ersten Mal seit fast zwei Jahren unter \textit{10 Millionen gefallen ist}.\\[1ex]
LLM (Azure OpenAI &	Ein Bericht des Arbeitsministeriums in dieser Woche zeigte auch, dass die Anzahl der verfügbaren Stellen  \\
with custom delimiters)&\textit{im Februar unter 10 Millionen fiel}, erstmals seit fast zwei Jahren.\\[1ex]
Hybrid	&Ein Bericht des Arbeitsministeriums in dieser Woche zeigte zudem, dass die Zahl der verfügbaren Stellen\\
&\textit{im Februar} erstmals seit fast zwei Jahren \textit{unter 10 Millionen fiel}. \\\hline

    \end{tabular}
    \caption{Examples from the qualitative comparison of the systems when translating English with text styling (italics) to German. Note that the German translations vary slightly between the systems.}
    \label{tab:exampleonetwothree}
\end{table*}

\subsection{Insights from Evaluations}
\label{sec:insights}

None of the NMT-based examples included hallucinations in the translations. Hallucinations can be a major issue in LLM-based approaches to translation and more generally (see for example \citep{dale-etal-2023-detecting,guerreiro-etal-2023-optimal}). In addition, none of the mistakes in style transfer showed bias or ethical issues.

In examples where the stylized text was short ($<$4 words) and the translations had contiguous elements, all technologies were on par with the attention head method.
All the technologies also work well when style needs to be transferred between named entities (e.g.\ names of people,  locations). These named entities form a contiguous phrase in both the target and source language.

Issues occurred in cases where the stylized text broke into two parts, i.e.\ were separated by other words. The NMT method had problems, either adding text styling on the intervening words or dropping styling on some words. Interestingly, the LLMs  generate translations which put all the words of the stylized phrase together by either changing the tone or the syntactic voice (active or passive) of the sentence. This causes subtle tone differences and does not match the designer’s expectations.	
Finally, when there are multiple occurrences of a word in the translated text,  these systems   apply the styling on the last occurrence of that word in the sentence, which  changes the style and aesthetic of the translated sentence.

\section{	Conclusions and Future Directions}

This paper examines techniques for transferring text styling from original source content to translated content, a key component of design applications serving global markets. Interestingly, the  strong baseline of using attention heads from NMT with a specialized scoring logic outperformed the three proposed techniques. The proposed techniques introduce  custom tags  with NMT input and LLM prompts to transfer the styles gracefully.  Transferring the styling with NMT produces high quality translation but sometimes transfers the styles inaccurately. The LLM-based solution  transfer the markers  more accurately, but their cost and latency is higher and their translation quality is poorer than the NMT approach. From an output quality perspective, the hybrid approach  brings the best of both the worlds with performant content translation via NMT and leveraging the LLM  to identifying the unigram mappings for style transfer. 
As  LLMs evolve, using the attention heads from the LLM processing layers and  leveraging specialized LLMs conditioned around translation  could lead to an architecture well suited for translations with text style transfer. 

As future work,  these techniques need to account for  culturalization while transferring the styling, e.g.\ whether the choice of font or color is acceptable to a particular region.  This is an orthogonal dimension to the translation and style transfer discussed here. We aim to explore how LLMs can be used  to address the text content tonality and cultural preferences in order to further improve design globalization. For the text tonality within the translation component, few-shot learning and prompting \citep{sanh2022multitask,vincent-etal-2023-mtcue} could help designers and enterprises to leverage  their existing translated documents to guide the system to honour similar tonality and language constructs. An initial step towards this it to provide more immediate context information to the LLM styling component such as document type (e.g.\ magazine cover, mobile advertisement, brochure) via the prompt. Overall, the broad capabilities of the LLMs combined with the high quality NMT translations suggest a bright future for improving the creation process for graphic designers.

%\balance
%\newpage
\bibliographystyle{ACM-Reference-Format}
\bibliography{main}

\end{document}